# [WIP] Medical Incident Prediction Through Analysis of Electronic Medical Records Using Machine Learning

~Fall Prediction~

Atsushi Yanagisawa[*1], Chintaka Premachandra[*2], Hiroharu Kawanaka[*3], Atsushi Inoue[*3], Takeo Hata[*4], Eiichiro Ueda[*4]

[*1]Takatsuki High School, Japan    [*2]Shibaura Institute of Technology, Japan

[*3]Mie University, Japan    [*4]Osaka Medical and Pharmaceutical University, Japan

E-mail: email@inoueatsushi.net, eiueda@ompu.ac.jp

*Abstract*—This work-in-progress paper reports our preliminary work on medical incident prediction in general, and fall risk prediction in specific, using machine learning. Data for the machine learning are generated only from the particular subset of the electronic medical records (EMR) at Osaka Medical and Pharmaceutical University Hospital. As a result of conducting three experiments such as (1) machine learning algorithm comparison, (2) handling imbalancement, and (3) investigation of explanatory variable contribution to the fall incident prediction, we find the investigation of explanatory variables the most effective.

*Contribution*—Work-in-progress on medical incident prediction using machine learning.

## I. Introduction.

Fall incidents occur in 30%-60% of older adults each year, and the 10%-25% of these incidents result in fractures[1]. Hip fractures are particularly detrimental to the daily activities of elderly people[2], which in turn increases the social burden by increasing the mortality rate and its corresponding economic cost. That leads to a lower Activities of Daily Living (ADL) and the corresponding Quality of Life (QOL)[3,4]. In fact, more than 90% of hip fractures occur because of falls[5].

Falls in a hospital may cause trauma and bone fracture, which may greatly affect both ADL and QOL[6], and may often lead to lawsuits and additional medical costs[7,8,9]. Falls are a major cause of mortality and morbidity in earlier adultings[10]. Fall prevention is therefore important to reduce fractures in elderly adults in order to allow them to continue to live autonomously[11].

For the above reasons, we kicked off a research project on medical incident prediction using machine learning. First, we are interested in the incident prediction for in-patients who fall in a hospital. Second, we are interested in whether or not it is possible to predict such fall incidents based only on their electronic medical records (EMR). Third, we are interested in the effectiveness and efficiency of machine learning tools such as Pytorch and scikit-learn[25] that are popular in critical practice and applications like ours.

## II. Background and Motivation

Early detection of risk factors, preferably before falls have occurred, would enable earlier intervention to reduce falls and subsequent mortality and morbidity[12]. Understanding factors associated with falls plays a vital role in preventing any falls in the future[13]. However, its ability to predict falls has not been fully investigated in-patients with multiple sclerosis (PwMS)[14]. The rate prevalence of falls in the present study was 13.5%, which is lower than that reported in previous studies[15,16,17]. In fact, the rate of fall incidents at Osaka Medical and Pharmaceutical University Hospital is likely as low due to their medical safety practice for in-patients.

It was not possible to decipher the potential differential effects of individual antipsychotic agents and their relationship with falls and fractures[18]. This study may mix up various risks whether high or low because we aggregate prescriptions of all psychotropic drugs into analytics. On the other hand, the number of particular prescriptions is too small for each of such drugs. Additionally, the number of incidents itself is significantly smaller (thus imbalance) as well.

When considering fall risk factors, various medications including psychotropic and antipsychotics, neuropsychiatric symptoms, cognitive metrics such as Mini-Mental State Examination scores, or functional problems did not predict hospitalized falls[19]. A more recent systematic review[20] confirmed the earlier risk factors and also found mixed results for sociodemographic risk factors, while noting that psychotropic medication, particularly benzodiazepines and antipsychotics, and orthostatic hypotension were associated with an increased risk of falls[19]. Several known risk factors for falls in previous work were not found to be significant predictors in this study. Women, older persons, those with higher body mass index, those with multiple arthroplasty procedures, and persons taking opioids were not at higher risk of falls[21].

We identify several limitations while collecting data for this study. First, our fall history data include self-reporting. Arguably the most serious limitation is the recall bias introduced by the self-administered questionnaires for evaluating comorbidities and fall histories[11]. We completely depended on diagnostic coding for the identification of coccyx fracture patients in this database study and there was a possibility of overestimation due to erroneous coding in the medical claims[22]. On the other hand, there may be disease names used for insurances. Furthermore, comorbidities we selected may include diseases that are already cured.

The longer period of fall history increases the exposure, which may explain the much larger incidence of falling. However, the longer time period in that study may inflate issues related to recall bias, and the small sample size may reduce the generalizability of those results[23]. Since fall history does not include the exact timestamp of those incidents, the duration at-risk may or may not be accurate and consistent. Although weak muscles are commonly thought to be related to fall risk, it may or may not be predictive in this analysis due to the same reason as pain. Patients with less activity may have greater weakness or disuse atrophy, thus they also have less exposure or opportunities to fall. Future work should consider active time or number of exposures to account for this confounding variable[23]. Because in-patients in more critical conditions have less opportunities to fall, there is a possibility that the fall rate appears to be lower than actual. Last but not the least, there is a concern that our follow-up duration may be too short because we only track their hospitalization periods.

### III. EXPERIMENTS AND STUDIES

Three experiments have been conducted so far. More experiments will follow in order to complete this study.

*A. Experiment 1. Model (binary classifier) generation using various machine learning algorithms.*

Objectives. To be able to generate models with the highest possible precision using various machine learning algorithms.

Hypothesis. Because our data set includes many categorical explanatory variables, we expect linear model based approaches not to be effective. On the contrary, discrete and case-based algorithms such as k-nearest neighbor and decision tree learning are expected to be effective to handle categorical variables.

Details.

*Machine learning algorithms.

1. Support Vector Machine with a linear kernel function (linear model).
2. Logistic regression.
3. Gradient Boosting Machine (Decision Tree). This generates a very sharow decision tree as the model and uses a minimal amount of data. This is known to be robust against overlearning.
4. k-nearest neighbor (with k=1,2,3,4).

*Dataset.

We extract data from the Electronic Medical Record (EMR) at Osaka Medical and Pharmaceutical University Hospital. In particular, the EMR for 103199 in-patients from 2014 to 2018 relevant to falls are extracted. This consists of 172 exploratory variables and the binary label of falling (or not), such as department, medication, diagnosis, and basic metrics of patients such as weight and age. Dummy coding has been applied as necessary to machine learning. This contains 1213 falls and 101986 no-falls (i.e. clearly imbalanced). We randomly select the same number of no-falls, i.e. 1213, in order to handle this imbalancement. Other than that, there are no adjustments applied on any variables such as normalization and regularization.

*Model evaluation.

We use the following metrics such as

- Accuracy: (TP+NP)/(TP+TN+FP+FN),
- Recall: TP/(TP+FN), and
- Precision (aka positive predictive value): TP/(TP+FP),

where

- TP: true positive,
- TN: true negative,
- FP: false positive, and
- FN: false negative.

We use accuracy as the overall performance indicator and recall as the fall sensitivity indicator.

*Procedure.

1. Using 'Random Under Sampling', we prepare the dataset for model generation -- 1213 records of each fall and no-fall. 2426 total. (Actually, randomly eliminating no-fall records until being equal to the number of falls.)
2. Devide randomly the data set into two: 90% of training data and 10% of test data.
3. Model generation using various machine learning algorithms.
4. Compute the accuracy for both training and test datasets and the recall for test dataset.

Study.

Table I. Results in experiment 1.

| ML Algorithm | Accuracy (training) | Accuracy (testing) | Recall |
|---|---|---|---|
| Support Vector Machine | 0.6483 | 0.6543 | 0.8376 |
| Logistic Regression | 0.7262 | 0.7242 | 0.6923 |
| Gradient Boosting Machine | 0.7893 | 0.8271 | 0.8291 |
| k-nearest neighbor | 0.7279 | 0.6831 | 0.6581 |

(Rounded at the fifth digit)

Table I shows the results. On the contrary to hypotheses, k-nearest neighbor does not outperform linear model approaches, and this does not even reach to 0.9 in its recall or accuracy. The number of exploratory variables is likely too high. As expected, Gradient Boosting Machine generally outperforms the others. As consistent with the theory, decision tree learning is effective to many exploratory variables.

On the other hand, the highest metrics value is around 0.83-0.84 overall. We need to study whether this is sufficient for the involved medical practice. Overall, there are questions on both quality and quantity of the dataset -- e.g., consistency, imbalancement, and high-dimensionality.

B. *Experiment 2. Handling the imbalancement in preprocess (dataset preparation).*

Objectives. To be able to prepare a dataset by handling the imbalancement.

Hypothesis. We expect to generate models with better performance as a result of handing imbalancement.

Details.

*Machine learning algorithms.

We use the same four algorithms as in experiment 1.

*Datasets.

From the same extracted data prepared in experiment 1, we generate three different datasets. Different from experiment 1, we first divide the fall data into two: training (90%) and test (10%). We then put together no-fall data with different preparations. The reason why we divide the fall data first is to avoid redundancies in the oversampling (generated) fall data.

- Dataset 1: Random Under Sampling. The same as experiment 1 -- undersampling (eliminating) no-fall data until the same number as the falls is left.
- Dataset 2: Random Over Sampling according to the statistics (mean and variance). Unlike experiment 1, we oversample fall data according to the statistics (mean and variance) until the same number as the no-falls is generated.
- Dataset 3: Synthetic Minority Over-sampling Technique (SMOTE). Unlike experiment 1, we oversample fall data based on its correlation until the same number as the no-falls is generated.

We use default parameters for all three dataset preparations.

*Model evaluation. We use the recall and the confusion matrix itself, i.e. consisting of TP, FP, TN, and FN, as indicators.

*Procedure.

1. Divide the fall data into two: training (90%) and test (10%).
2. Generate three datasets as described above.
3. Generate models using various machine learning algorithms for each dataset.
4. Using test data for each dataset, we obtain indicators.

Study.

Tables II, III and IV show the results. Different from the hypothesis, we did not obtain higher recalls as a result of oversampling data. The linear models (SVM and Logistic Regression) do not make any significant difference in recalls and confusion matrix for undersampling and oversampling. This is an indication of those linear models accommodating this complex data. On the other hand, Gradient Boost Model and k-nearest neighbor do not make better recalls for oversampling. This is consistent with the previous finding of no accommodation for highly-dimensional, sparse data.

Table II. Results in experiment 2: Random Undersampling no-fall-over data.

| ML Algorithm | Recall | TN | FP | FN | TP |
|---|---|---|---|---|---|
| Support Vector Machine | 0.7364 | 7236 | 2955 | 34 | 95 |
| Logistic Regression | 0.7442 | 7245 | 2946 | 33 | 96 |
| Gradient Boost Machine | 0.7597 | 7138 | 3053 | 31 | 98 |
| K-nearest neighbor | 0.6667 | 6903 | 3288 | 43 | 86 |
| Mean | 0.7264 | 7130.5 | 3060.5 | 35.2 | 93.7 |

(Rounded at the fifth digit)

Table III. Results in experiment 2: Random Oversampling consistent with mean and variance.

| ML Algorithm | Recall | TN | FP | FN | TP |
|---|---|---|---|---|---|
| Support Vector Machine | 0.7132 | 7372 | 2819 | 37 | 92 |
| Logistic Regression | 0.7287 | 7412 | 2779 | 35 | 94 |
| Gradient Boost Machine | 0.6357 | 7955 | 2236 | 47 | 82 |
| K-nearest neighbor | 0.0775 | 9898 | 293 | 119 | 10 |
| Mean | 0.5387 | 8159 | 2031 | 59.5 | 69.5 |

(Rounded at the fifth digit)

Table IV. Results in experiment 2: SMOTE -- Random Oversampling based on the correlation.

| ML Algorithm | Recall | TN | FP | FN | TP |
|---|---|---|---|---|---|
| Support Vector Machine | 0.7132 | 7463 | 2728 | 37 | 92 |
| Logistic Regression | 0.7287 | 7504 | 2687 | 35 | 94 |
| Gradient Boost Machine | 0 | 10191 | 0 | 129 | 0 |
| K-nearest neighbor | 0.1705 | 9525 | 666 | 107 | 22 |
| Mean | 0.4031 | 8670 | 1520 | 77 | 52 |

(Rounded at the fifth digit)

There is no difference in the performance of linear models between cases of undersampling and oversampling. However, k-nearest neighbor makes a significant difference in recalls -- 0.0542. This must have something to do with its power of generalization such that k-nearest neighbor works well only when the input for prediction is close enough to the collected samples.

Handling imbalancement by both oversampling and undersampling does not improve the prediction performance of models. This suggests that only several selective explanatory variables (in other words, a simpler and more intuitive model) should be used per Occam's Razor. Prelude to this, we think of investigating models generated by only one explanatory variable.

C. *Experiment 3. Study on exploratory variables.*

Objectives. To be able to know the contribution of a single explanatory variable to the prediction performance.

Hypothesis. Explanatory variables such as diabetes, malignant tumor, peripheral neuropathy, and circulatory disorders that are thought to be more relevant to falls should contribute more to the prediction performance.

Details.

*Machine learning algorithms. We use the same as in experiment 1.

*Dataset. We use the same dataset as in experiment 2.

*Model evaluation.

We use recall and precision as the performance indicators. We use recall because the numbers of falls and no-falls are not necessarily the same in test data. The precision is affected by such an unequal ratio, but the evaluation only based on recall is not trustworthy enough. For example, we cannot be certain for a fall predicted even if its recall is 1 (i.e. no false negative). In fact, the prediction may be a false positive.

*Procedure.

1. Extract one explanatory variable from the dataset.
2. Randomly divide the extracted data in to two: for training (90%) and for test (10%).
3. Generate the training data.
4. Generate models for various machine learning algorithms.
5. Compute recall, precision, correlation; as well as, mean and median for all, falls and no-falls.

We perform this procedure for each explanatory variable. We only perform this for several selective ones due to the time constraints.

Study.

Overall, explanatory variables about or related to patient age, operation, long-term hospitalization, and department are found 'highly involved' for the fall prediction. Table V shows the details. Unlike the hypothesis, those about diabetes and peripheral nerve disorder did not contribute significantly enough to generate the satisfactory prediction models. Many of them did not satisfy precision (i.e. too many false positives). The reason why the model created from the above explanatory variables had a high reproduction value can be considered medically and statistically as follows.

*Patient age and #mo.

From medical aspects, physical and cognitive functionalities decrease. Fall incidents are considered to be caused because of that. The prediction models based on such a variable are consistent with that as well, and the recall is high as a result. Moreover, the differences of means and medians between falls and no-falls are 11.5 and 8 respectively. Those differences are more significant than those of patient heights, i.e. 4 and 0.5 respectively. This leads to relatively clear value clusters for falls and no-falls.

*Gynecology.

Many patients in the gynecology department receive chemotherapy due to gynecologic cancer. This leads to various adverse drug reactions more so than other medications. Such severe reactions often serve as a trigger of lower mortality, thus resulting in fall incidents. In addition, the difference of means is 0.0778, that is much larger than that for perceptual impairment (i.e. 0.03). Therefore, this is quite clearly clustered for falls and no-falls among all categorical explanatory variables.

*Cardiology.

Patients in Cardiology department often receive antihypertensive medications that causes lightheadedness and dizziness as a result of their lower blood pressure. Then, such lightheadedness and dizziness are the cause of falls. Unfortunately, the difference of means of this explanatory variable is similar to that for vision impairment, and no significance is found for the fall prediction.

*Ophthalmology

It is thought that vision problems such as monocular vision with poor eyesight make it difficult to recognize things and make falls more likely. Furthermore, quite clear clusters for falls and no-falls have been observed, which is similar to the case of Gynecology.

*Anesthesia period and Operation period.

The longer these times are, the more likely it is to adversely affect the physical condition after surgery. In addition, anesthesia itself carries the risk of various complications, which can lead to fall incidents. Furthermore, quite clear clusters for falls and no-falls have been observed, too.

*Plan A-1.

Patients under Plan A-1 has the highest ADL (i.e. activity of daily life), thus they are allowed to move freely within the hospital. Unfortunately, such freedom may result in increasing the risk of falls. In fact, quite clear clusters for falls and no-falls have been observed. However, our current dataset for this project only contains the explanatory variable to indicate only the Plan A-1 or not but no relevant measures. In fact, the hospital has an oral alert practice for the A-1 in-patients with fall potentials, that should suppress the risk.

*Autonomy, check-up purpose, and planned hospitalization.

Similar to Plan A-1. However, quite significant clear clusters for falls and no-falls have been observed on explanatory variables for autonomy and check-up purpose hospitalization, while no such clusters have been observed on the explanatory variable for planned hospitalization.

*AAA medication (Antipyretic Analgesic and Anti-inflammatory).

Since this kind of medication is applied when feavering or to relieve pains after a surgery, the fall risk becomes significant during such a symptom is observed. Unfortunately, the difference of means of this explanatory variable is similar to that for vision impairment, and no significance is found for the fall prediction.

IV. RESPONSES FROM MEDICAL EXPERTS

After conducting experiment 3, those variables mentioned above are found to be significantly involved in the higher (but not necessarily satisfactory) prediction performance. There are two responses from medical experts obtained.

Response 1: The following five factors out of the ten explanatory variables are generally consistent with previous reports: older age, cardiovascular disease, patients undergoing cancer chemotherapy (internal medicine), longer anesthesia times, and longer surgery times. There are unlikely no studies reported that directly find direct causes as far as I know. Medical experts may reasonably accept those 10 variables as fall risk factors. Older people often have lower physical functions such as muscle strength and balance, cognitive decline (difficult to detect dangers such as steps), and cardiovascular diseases. Although not directly, taking antihypertensive drugs lowers blood pressure. That is considered a fall risk. When blood pressure decreases, it causes light-headedness and lightheadedness (orthostatic hypotension), which poses another fall risk. Patient's general condition deteriorates while receiving cancer chemotherapy because severe side effects may likely occur more so than other medications -- nausea, fever, dizziness, peripheral neuropathy, liver damage, blood damage, kidney damage, heart damage, and immune-related side effects called irAE in some severe cases. Such physical conditions may result in yet another high fall risk. Longer period of anesthesia and operations resulting in yet another high fall risk is considered appropriate as there is more invasiveness that affects a patient's conditions after surgery. General anesthesia itself carries the risk of various complications. Therefore, the longer period of anesthesia and operation resulting in yet another high fall risk is considered appropriate as well. Other variables likely seem confounding. For instance, gynecology is likely confounded with chemotherapy, and other factors must be confounded for the planned and planned hospitalization themselves. Patients with Plan A-1 is an ideal case of nursing. On the contrary, the fall risk may be high as a result of their free movement.

<u>Response 2</u>: AAA medication (Antipyretic Analgesic and Anti-inflammatory) itself is hard to serve as a high fall risk factor, is thus likely cofundedness. The symptoms of fevering, pain, etc. themselves affect the fall risk, too. Such circumstances with symptoms and medications are naturally considered as high fall risks. Indeed, there are reports of this kind such as [26].

## V. CONCLUDING REMARKS

Our work-in-progress has been reported. Three experiments were conducted, and concluded that investigating the contribution of explanatory variables for the fall risk prediction was the most effective.

As our experiments are still rough, thus preliminary, we need to further conduct investigations on the contribution of explanatory variables and their combinations for a better prediction performance. Afterwards, we will be ready to implement a medical risk prediction system that works integratively with various medical information systems including but not necessarily limited to EMR, nursing reports, billing, medications, and others.

## ACKNOWLEDGMENT

We would like to acknowledge and to express our sincere gratitude to the following faculty members at Takatsuki High School, Japan: Toru Ohki, Aya Ochida and Yuko Nagao. They fully support this comprehensive machine learning work conducted by Atsushi Yanagisawa as a part of their curricular activities related to the Super Science High School Program. This project would not make progress without their support.

This work is financially supported by Speedy, Inc. (http://spdy.jp/) We would like to acknowledge and to express our sincere gratitude to Atsushi Fukuda, the President and CEO of Speedy, Inc.

## REFERENCES

[1] Rubenstein LZ. Falls in older people: epidemiology, risk factors and strategies for prevention. Age Ageing, 35 Suppl 2, ii37–ii41 (2006).

[2] Theander E, Jarnlo G-B, Ornstein E, Karlsson M. Activities of daily living decrease similarly in hospital-treated patients with a hip fracture or a vertebral fracture: a one-year prospective study in 151 patients. Scand. J. Public Health, 32, 356–60 (2004).

[3] Hagino H. Fragility fracture prevention: review from a Japanese perspective. Yonago Acta Med., 55, 21–8 (2012).

[4] Ohta H, Mouri M, Kuroda T, Nakamura T, Shiraki M, Orimo H. Decreased rate of hip fracture and consequent reduction in estimated medical costs in Japan. J. Bone Miner. Metab., 35, 351–353 (2017).

[5] Parkkari J, Kannus P, Palvanen M, Natri A, Vainio J, Aho H, Vuori I, Järvinen M. Majority of hip fractures occur as a result of a fall and impact on the greater trochanter of the femur: a prospective controlled hip fracture study with 206 consecutive patients. Calcif. Tissue Int., 65, 183–7 (1999).

[6] Imagama S, Ito Z, Wakao N, Seki T, Hirano K, Muramoto A, Sakai Y, Matsuyama Y, Hamajima N, Ishiguro N, Hasegawa Y. Influence of spinal sagittal alignment, body balance, muscle strength, and physical ability on falling of middle-aged and elderly males. Eur. Spine J., 22, 1346–53 (2013).

[7] van Weel C, Vermeulen H, van den Bosch W. Falls, a community care perspective. Lancet (London, England), 345, 1549–51 (1995).

[8] Healey F, Scobie S, Oliver D, Pryce A, Thomson R, Glampson B. Falls in English and Welsh hospitals: a national observational study based on retrospective analysis of 12 months of patient safety incident reports. Qual. Saf. Health Care, 17, 424–30 (2008).

[9] Udén G. Inpatient accidents in hospitals. J. Am. Geriatr. Soc., 33, 833–41 (1985).

[10] Evans D, Pester J, Vera L, Jeanmonod D, Jeanmonod R. Elderly fall patients triaged to the trauma bay: age, injury patterns, and mortality risk. Am. J. Emerg. Med., 33, 1635–8 (2015).

[11] Arai T, Fujita H, Maruya K, Morita Y, Asahi R, Ishibashi H. The one-leg portion of the Stand-Up Test predicts fall risk in aged individuals: A prospective cohort study. J. Orthop. Sci., 25, 688–692 (2020).

[12] Best JR, Davis JC, Liu-Ambrose T. Longitudinal Analysis of Physical Performance, Functional Status, Physical Activity, and Mood in Relation to Executive Function in Older Adults Who Fall. J. Am. Geriatr. Soc., 63, 1112–20 (2015).

[13] Vu HM, Nguyen LH, Nguyen HLT, Vu GT, Nguyen CT, Hoang TN, Tran TH, Pham KTH, A Latkin C, Xuan Tran B, S H Ho C, Ho RCM. Individual and Environmental Factors Associated with Recurrent Falls in Elderly Patients Hospitalized after Falls. Int. J. Environ. Res. Public Health, 17, 2441 (2020).

[14] Tajali S, Mehrvar M, Negahban H, van Dieën JH, Shaterzadeh-Yazdi M-J, Mofateh R. Impaired local dynamic stability during treadmill walking predicts future falls in patients with multiple sclerosis: A prospective cohort study. Clin. Biomech. (Bristol, Avon), 67, 197–201 (2019).

[15] Muraki S, Akune T, Ishimoto Y, Nagata K, Yoshida M, Tanaka S, Oka H, Kawaguchi H, Nakamura K, Yoshimura N. Risk factors for falls in a longitudinal population-based cohort study of Japanese men and women: the ROAD Study. Bone, 52, 516–23 (2013).

[16] Muraki S, Akune T, Oka H, Ishimoto Y, Nagata K, Yoshida M, Tokimura F, Nakamura K, Kawaguchi H, Yoshimura N. Physical performance, bone and joint diseases, and incidence of falls in Japanese men and women: a longitudinal cohort study. Osteoporos. Int., 24, 459–66 (2013).

[17] Matsumoto H, Tanimura C, Tanishima S, Osaki M, Noma H, Hagino H. Sarcopenia is a risk factor for falling in independently living Japanese older adults: A 2-year prospective cohort study of the GAINA study. Geriatr. Gerontol. Int., 17, 2124–2130 (2017).

[18] Stubbs B, Mueller C, Gaughran F, Lally J, Vancampfort D, Lamb SE, Koyanagi A, Sharma S, Stewart R, Perera G. Predictors of falls and fractures leading to hospitalization in people with schizophrenia spectrum disorder: A large representative cohort study. Schizophr. Res., 201, 70–78 (2018).

[19] Sharma S, Mueller C, Stewart R, Veronese N, Vancampfort D, Koyanagi A, Lamb SE, Perera G, Stubbs B. Predictors of Falls and Fractures Leading to Hospitalization in People With Dementia: A Representative Cohort Study. J. Am. Med. Dir. Assoc., 19, 607–612 (2018).

[20] Fernando E, Fraser M, Hendriksen J, Kim CH, Muir-Hunter SW. Risk Factors Associated with Falls in Older Adults with Dementia: A Systematic Review. Physiother. Can., 69, 161–170 (2017).

[21] Riddle DL, Golladay GJ. Preoperative Risk Factors for Postoperative Falls in Persons Undergoing Hip or Knee Arthroplasty: A Longitudinal Study of Data From the Osteoarthritis Initiative. Arch. Phys. Med. Rehabil., 99, 967–972 (2018).

[22] Won H, Moon S-Y, Park JH, Kim J-K, Kim HS, Baek S-H, Kim S-Y, Lee Y-K, Koo K-H. Epidemiology and risk factors of coccyx fracture: A study using national claim database in South Korea. Injury, 51, 2278–2282 (2020).

[23] Aljehani MS, Crenshaw JR, Rubano JJ, Dellose SM, Zeni JA. Falling risk in patients with end-stage knee osteoarthritis. Clin. Rheumatol., 40, 3–9 (2021).

[24] https://pubmed.ncbi.nlm.nih.gov/2017229/

[25] Andreas C. Muller and Sarah Guido, Introduction to Machine Learning with Python, OREILLY.

[26] T. Suzuki, Epidemiology of fall-over, Journal of Japanese Eldery Medical Society 40, pp. 85-94.

Table V. Results in experiment 3.
(We select some explanatory variables with recall >0.8 and precision >0.013. Prediction performance with recall higher than its mean tends to be consistent. Moreover, the precision becomes 0.01-0.012 when the model predicts fall-over for all inputs from the test data -- i.e. no prediction capacity. Under such circumstances, we set a guideline with respect to the recall and the precision for models that use a single explanatory variable.)

| Explanatory Variable | Recall | Prec. | Corr. | Mean: all | Median: all | Mean: fo | Median: fo | Mean: no fo | Median: no fo |
|---|---|---|---|---|---|---|---|---|---|
| Patient age | 0.9147 | 0.0155 | 0.0519 | 56.9 | 65 | 68.3 | 73 | 56.8 | 65 |
| Patient #mo | 0.9457 | 0.0138 | 0.0519 | 689.1 | 790 | 825.4 | 880 | 687.5 | 789 |
| Gynecology | 0.9767 | 0.0135 | -0.0279 | 0.0988 | 0 | 0.022 | 0 | 0.0998 | 0 |
| Cardiology | 0.9612 | 0.0130 | -0.0065 | 0.0806 | 0 | 0.0643 | 0 | 0.0808 | 0 |
| Ophthalmology | 0.9922 | 0.0135 | -0.0268 | 0.0865 | 0 | 0.0173 | 0 | 0.0873 | 0 |
| Anesthesia period: > average | 0.9224 | 0.0136 | -0.0211 | 0.1480 | 0 | 0.0791 | 0 | 0.1489 | 0 |
| Operation period: > average | 0.9147 | 0.0142 | -0.0234 | 0.1959 | 0 | 0.1104 | 0 | 0.1969 | 0 |
| Plan A-1 | 0.9379 | 0.0158 | -0.044 | 0.2615 | 0 | 0.0832 | 0 | 0.2636 | 0 |
| Autonomy | 0.9534 | 0.0158 | -0.042 | 0.2497 | 0 | 0.0799 | 0 | 0.2517 | 0 |
| AAA* medication | 0.9612 | 0.0130 | -0.004 | 0.0798 | 0 | 0.0676 | 0 | 0.0800 | 0 |
| Check-up purpose | 0.9379 | 0.0130 | -0.019 | 0.1106 | 0 | 0.0535 | 0 | 0.1100 | 0 |
| ER:Planned | 0.9844 | 0.0131 | -0.006 | 0.0675 | 0 | 0.0519 | 0 | 0.0677 | 0 |
| (Ref.) Vis. impair. | - | - | 0.008 | 0.2135 | 0 | 0.2464 | 0 | 0.2131 | 0 |
| (Ref.) height | - | - | 0.012 | 152 | 158.5 | 156.0 | 159 | 152.7 | 158.5 |

(*AAA: Antipyretic Analgesic and Anti-inflammatory. The highest recall and precision among all models are selected. Rounded at the fifth digit. Plan A-1 designates in-patients whose autonomy is normal. All variables except patient age, patient #mo and height are categorical. Visual impairment and height are listed as references.)